\documentclass[letterpaper, 10 pt, conference]{ieeeconf}  
\IEEEoverridecommandlockouts                              
\usepackage{soul}
\usepackage{url}
\usepackage[hidelinks]{hyperref}
\usepackage[utf8]{inputenc}
\usepackage[small]{caption}
\usepackage{graphicx}
\usepackage{amsmath}
\usepackage{booktabs}
\usepackage[T1]{fontenc}
\usepackage{amssymb,amsfonts}
\usepackage{textcomp}
\usepackage{lineno}
\usepackage{subfigure}
\usepackage{float}

\usepackage{helvet}
\usepackage{mathptmx}

\usepackage{tabularx}
\usepackage{fancyhdr}
\usepackage{algorithm}    
\usepackage{algorithmic} 

\usepackage{epstopdf} %
\usepackage{color}
\usepackage{xspace}
\usepackage{graphicx}
\newcommand{\subparagraph}{}
\usepackage{titlesec}

\usepackage[dvipsnames]{xcolor}
\definecolor{mycolor0}{rgb}{0,0.4470,0.7410}
\definecolor{mycolor1}{rgb}{0.8500,0.3250,0.0980}
\definecolor{mycolor2}{rgb}{0.9290,0.6940,0.1250}
\definecolor{mycolor3}{rgb}{0.4940,0.1840,0.5560}
\definecolor{mycolor4}{rgb}{0.4660,0.6740,0.1880}
\definecolor{mycolor5}{rgb}{0.3010,0.7450,0.9330}
\definecolor{mycolor6}{rgb}{0.6350,0.0780,0.1840}
\colorlet{best}{ForestGreen}
\definecolor{SoftCyan}{RGB}{72, 202, 228}
\definecolor{VividCrimson}{RGB}{220, 20, 60}
\definecolor{DeepMagenta}{RGB}{139, 0, 139}
\definecolor{ForestGreen}{RGB}{189, 230, 205}
\definecolor{RoyalBlue}{RGB}{65, 105, 225}
\definecolor{BurntOrange}{RGB}{255, 217, 178}
\definecolor{MustardYellow}{RGB}{255, 245, 178}
\definecolor{Plum}{RGB}{142, 69, 133}
\definecolor{LightSeaGreen}{RGB}{32, 178, 170}
\definecolor{SlateGray}{RGB}{112, 128, 144}

\usepackage{colortbl}
\usepackage{graphicx}
\usepackage{tikz}
\usetikzlibrary{spy}
\usetikzlibrary{positioning}
\usetikzlibrary{shapes.geometric}
\usetikzlibrary{arrows.meta}
\usetikzlibrary{calc,angles}
\usetikzlibrary[shadings]
\tikzset{>={Stealth[width=3mm]}}
\usepackage{float}
\usepackage[binary-units,detect-all]{siunitx}
\usepackage{booktabs}
\usepackage{multirow}
\usepackage{threeparttable}
\usepackage{pgfplotstable}
\usepackage{pgfplots}
\usepackage{balance} 
\usepackage{bchart}
\usepgfplotslibrary{units}
\usepackage{xurl}
\usepackage{array}
\usepackage{calc}
\usepackage{csquotes}
\usepackage{tabularx}
\usepackage{etoolbox}
\usepackage{makecell}
\usepackage{colortbl}
\usepackage{stackengine}

\usepackage{setspace}
\renewcommand{\baselinestretch}{0.94}

\usepackage{comment}
\usepackage{cite}
\usepackage{amsmath,amssymb,amsfonts}
\usepackage{xcolor}
\def\BibTeX{{\rm B\kern-.05em{\sc i\kern-.025em b}\kern-.08em
    T\kern-.1667em\lower.7ex\hbox{E}\kern-.125emX}}

\begin{document}

\title{\LARGE\bf%
Gassidy: Gaussian Splatting SLAM in Dynamic Environments
}

\author{Long Wen$^{1}$, Shixin Li$^{1}$, Yu Zhang$^{1}$, Yuhong Huang$^{1}$, Jianjie Lin$^{1}$,\\ Fengjunjie Pan$^{1}$, Zhenshan Bing$^{1}$, and Alois Knoll$^{1}$
    \thanks{$^{1}$Authors from the Technical University of Munich, Munich, Germany. \{wenl, zha1, jianjie.lin, panf, knoll\}@in.tum.de}%
}

\maketitle
\thispagestyle{empty}
\pagestyle{empty}

\begin{abstract}
3D Gaussian Splatting (3DGS) allows flexible adjustments to scene representation, enabling continuous optimization of scene quality during dense visual simultaneous localization and mapping (SLAM) in static environments. 
However, 3DGS faces challenges in handling environmental disturbances from dynamic objects with irregular movement, leading to degradation in both camera tracking accuracy and map reconstruction quality. To address this challenge, we develop an RGB-D dense SLAM which is called Gaussian Splatting SLAM in Dynamic Environments (Gassidy). This approach calculates Gaussians to generate rendering loss flows for each environmental component based on a designed photometric-geometric loss function. To distinguish and filter environmental disturbances, we iteratively analyze rendering loss flows to detect features characterized by changes in loss values between dynamic objects and static components. This process ensures a clean environment for accurate scene reconstruction. Compared to state-of-the-art SLAM methods, experimental results on open datasets show that Gassidy improves camera tracking precision by up to 97.9\% and enhances map quality by up to 6\%. Video of experiments is available here: {\hypersetup{urlcolor=gray}\href{https://wen950223.wixsite.com/gassidy-slam-1}{https://www.wixsite.com.com/wen-Gassidy}}.

\end{abstract}

\section{Introduction}

Dense visual simultaneous localization and mapping (SLAM), known for its ability to present complex environments, is commonly used in tasks like mobile robot navigation~\cite{schops2019bad,dai2017bundlefusion,8421015}. These methods rely on known information about static environments to construct accurate maps. However, mobile robots often work in dynamic environments where unpredictable changes can reduce SLAM's mapping accuracy \cite{rosinol2023probabilistic,kahler2016real}. Therefore, addressing the challenges posed by dynamic environmental changes is essential to improving the effectiveness of SLAM in mobile robot tasks.

Recently, researchers have integrated Neural Radiance Fields (NeRF) into SLAM to reconstruct scenes in dynamic environments, as it can capture complex lighting effects and fine surface details \cite{Johari_2023_CVPR,Wang_2023_CVPR,NICE-SLAM}. 
Through generating optical flow and employing semantic segmentation, NeRF-based methods excel at filtering out disturbances of the challenging dynamic environment \cite{jiang2024rodyn,ONek}. 
However, these methods rely on predefined semantic segmentation to account for dynamic changes, which often fail to capture the irregular movements of objects \cite{jiang2024rodyn,ONek}.

3D Gaussian Splatting (3DGS), which constructs Gaussians independently to represent different regions of the scene, has emerged as a promising solution for this issue \cite{kerbl20233d}.  
This approach allows environmental changes to be flexibly expressed as changes of Gaussians within specific regions, relieving the need for predefined semantic masks \cite{sabour2024spotlesssplats,gao2024gaussianflow,lu2024gagaussian}. 
Despite its advantages, this approach is primarily tailored to handle photometric and geometric changes in static environments and faces challenges in accurately capturing dynamic objects within the scene.

\begin{figure}[t]
	\centering
    \subfigure[depth (ours)]{
    \label{comparison-1}
    \includegraphics[width=0.15\textwidth]{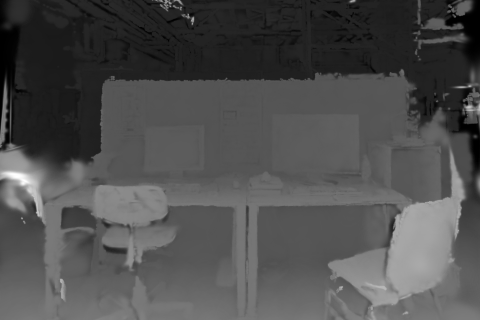}}%
    \hfill%
    \subfigure[Gaussian (ours)]{
    \label{comparison-2}
    \includegraphics[width=0.15\textwidth]{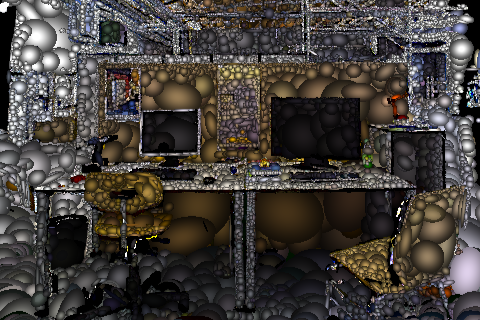}}%
    \hfill%
    \subfigure[RGB (ours)]{
    \label{comparison-3}
    \includegraphics[width=0.15\textwidth]{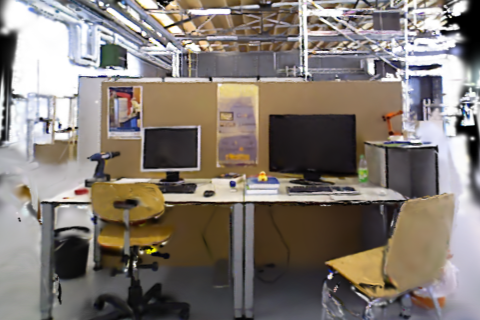}}
    
    \hspace{0.09em}
    \subfigure[depth (GSS)]{
    \label{comparison-4}
    \includegraphics[width=0.15\textwidth]{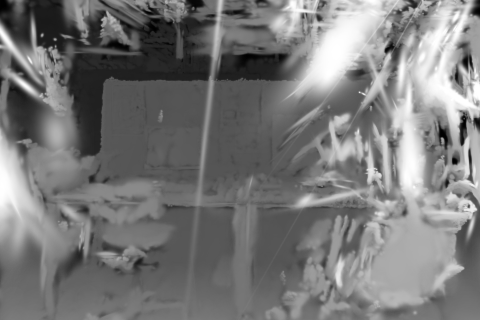}}%
    \hfill%
    \subfigure[Gaussian (GSS)]{
    \label{comparison-5}
    \includegraphics[width=0.15\textwidth]{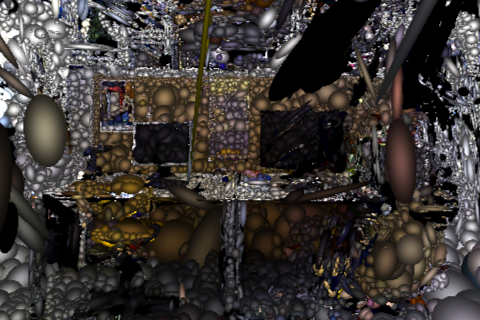}}%
    \hfill%
    \subfigure[RGB (GSS)]{
    \label{comparison-6}
    \includegraphics[width=0.15\textwidth]{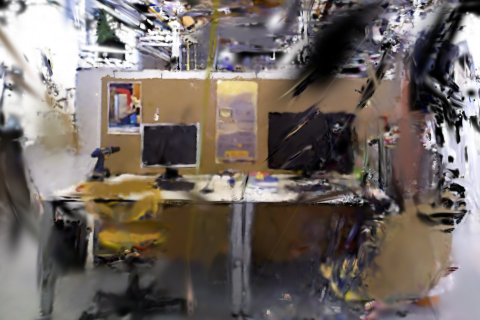}}%
    \caption{An example to illustrate the performance of Gassidy when compared with GS-SLAM (here GSS) \cite{Matsuki:Murai:etal:CVPR2024} on the TUM RGB-D dataset in the \textit{fr3/walk\_st} scene. The three images in a row represent the rendered depth, the created Gaussians, and the rendered RGB.}
	\label{comparison}
    \vspace{-1.5em}
\end{figure}
 
All the aforementioned works focus on recognizing environmental features to reconstruct the scene. 
However, such approaches either only consider static environments or struggle to effectively handle dynamic environmental changes. The reasons are multi-fold. First, disturbance from dynamic objects often causes overfitting during scene reconstruction, decreasing the accuracy of SLAM. 
Second, the change in dynamic objects is unpredictable, limiting the application of semantics that rely on prior dynamic knowledge. 
Third, dynamic objects with minor variations may be incorrectly identified as static environmental components in the viewpoint of subtle movements.

To address these challenges, this paper proposes an optimized 3DGS-based SLAM approach that incorporates rendering loss flows to analyze dynamic environments. We name this method \textbf{Ga}ussian \textbf{S}platting \textbf{S}LAM \textbf{i}n \textbf{Dy}namic environments (Gassidy). The approach is designed to filter out disturbances from dynamic objects while tracking the camera pose and reconstructing the scene. An example result of Mapping is illustrated in Fig.~\ref{comparison}. 

\begin{figure*}[t]
	\centering
	\includegraphics[width=0.98\textwidth]{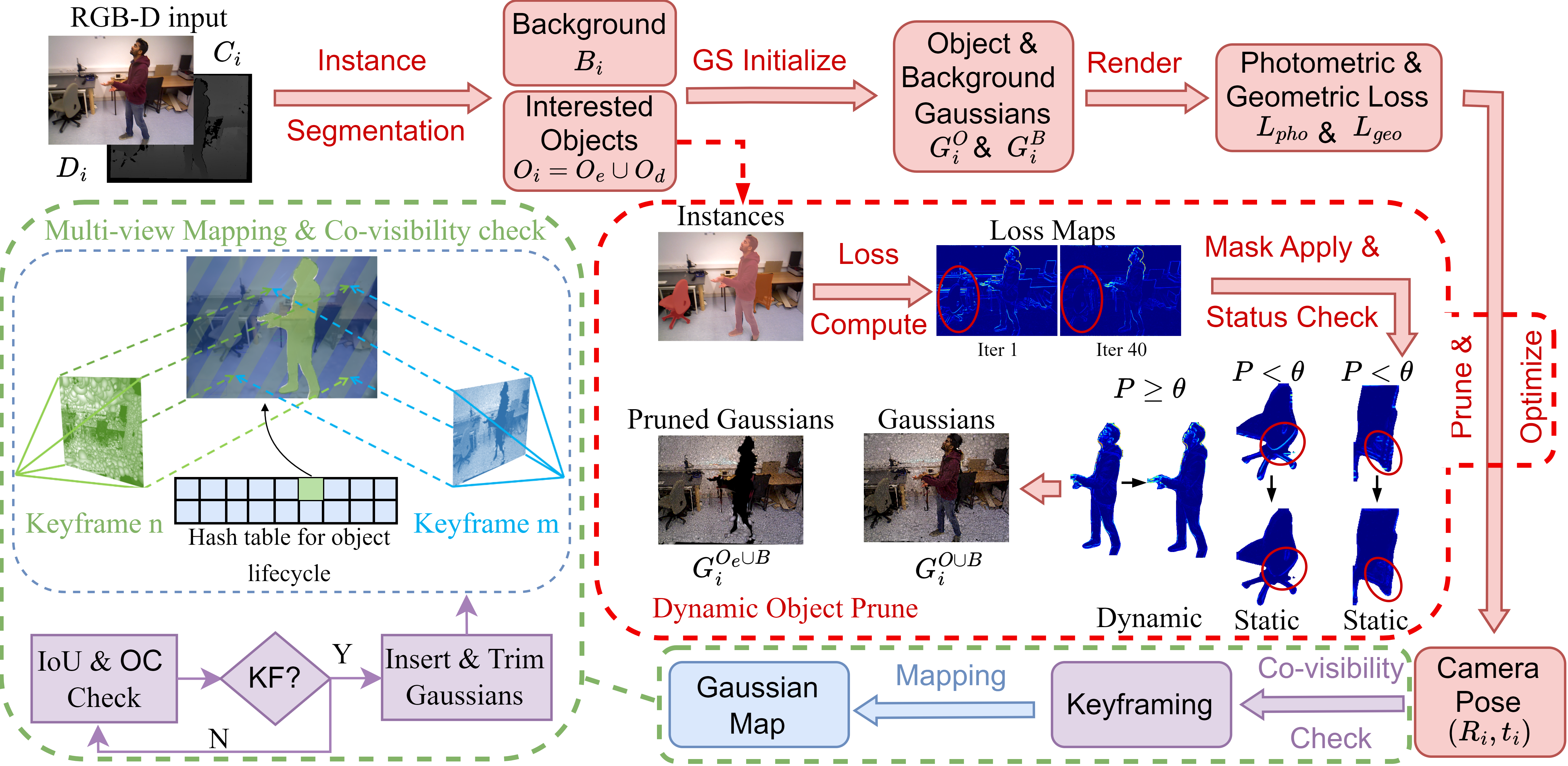}
    \caption{Architure of Gassidy: $i$ represents the frame number, $C_{i}$ and $D_{i}$ are the RGB image and corresponding aligned depth maps, $O_{i}$ and $B_{i}$ are the RGB-D sets of objects and background, $G_i^{O}$ and $G_i^{B}$ are the Gaussians of objects and background $L_{pho}$ and $L_{geo}$ are the photometric and geometric rendering loss. $R_i$ and $t_i$ are the rotation and translation parts of the camera pose. IoU and OC stand for Intersection over Union and Overlap Coefficient. $G^{O}$ and $G^{O_e}$ denote Gaussians for all objects and static objects, while $G^{B}$ represents background Gaussians. $P$ indicates the probability of an object being dynamic, and $\theta$ is the threshold.}
	\label{fig:workflow}
	\vspace{-1em}
\end{figure*} 

Our contributions are summarized as follows: 
\begin{itemize}
    \item To address unpredictable disturbances from dynamic objects, we use Gaussians to separately cover objects and background features, guided by instance segmentation. Environmental changes caused by dynamic objects are represented as variations in Gaussians rather than relying on predefined semantics.
    \item To distinguish between Gaussian feature changes caused by dynamic objects and those resulting from photometric or geometric variations in static environments, Gassidy calculates rendering loss flows for the Gaussians based on a designed photometric-geometric loss function to capture dynamic changes.
    \item To prevent misidentification of subtle object movements, the rendering loss flows are iteratively calculated to update the camera pose, amplifying the distinction between tiny object changes and subtle frame movements by analyzing features in loss value changes, thereby clearly identifying and filtering dynamic objects.
    \item Compared to state-of-the-art dense SLAM approaches, Gassidy achieves higher tracking precision of camera pose and reconstructs scenes with finer mapping quality when using widely-used open datasets (\enquote{TUM RGB-D} and \enquote{BONN Dynamic RGB-D}). In particular, when applying Gassidy, the tracking precision and mapping quality can be enhanced by up to 97.9\% and 6\%. 
\end{itemize}

\section{Related Work}
In recent years, the development of autonomous driving and robotics has necessitated detailed environmental maps, which are typically generated using dense visual SLAM methods.
To produce such maps, researchers always assume that the working environment is fixed and does not undergo significant changes during the task.
Consequently, SLAM methods generate scenes by treating the environment as static. 
For example, given fully known environmental information, SplaTAM directly optimizes geometric and photometric metrics based on 3D Gaussians to construct the target scene with explicit spatial extent \cite{keetha2024splatam}.
Similarly, GS-SLAM applies 3DGS to monocular SLAM, utilizing only RGB information to represent environmental features \cite{Matsuki:Murai:etal:CVPR2024}.
These methods can precisely reconstruct the scene by analyzing the detailed features of each frame.
However, in environments containing dynamic objects, these methods misidentify the dynamic objects as static, leading to overfitting in scene reconstruction and decreasing the mapping quality.

To handle dynamic environments, researchers aim to filter out dynamic objects by analyzing their movements in detail.
Khronos uses spatio-temporal methods to detect dynamic changes utilizing odometry input from other sensors \cite{Schmid2024Khronos}.
ONeK-SLAM combines feature points with NeRF to enhance object-level localization and reconstruction in environments with dynamic objects and varying illumination \cite{ONek}. 
RoDyn-SLAM enhances dense RGB-D SLAM in dynamic environments through a motion mask generation method and a divide-and-conquer pose optimization algorithm \cite{jiang2024rodyn}. 
By predefining the dynamic features of objects, these SLAM approaches can detect and filter out moving objects in dynamic environments, resulting in constructed scenes that align with the target ground truth.
However, because these methods rely on predefined semantic priors to model dynamic changes, they often fail to capture irregular or unpredictable object movements.

To address this issue, we introduce a novel dense SLAM method based on 3DGS that achieves high-quality mapping in dynamic environments by focusing on variations in the features of constructed Gaussians to distinguish dynamic objects, rather than relying on predefined knowledge.

\section{Designing Gassidy}

This section introduces Gassidy's architecture, details environmental feature extraction using 3DGS, explains filtering dynamic changes via rendering loss flow in tracking, and describes scene optimization with filtered Gaussians.

\subsection{Overview of the Gassidy}
\label{3a}

With the architecture of Gassidy illustrated in Fig.~\ref{fig:workflow}, our objective is to track the camera pose $(\mathrm{R}_i, \mathrm{t}_i)$ and generate a clean Gaussian map for scene reconstruction using the input image $\mathrm{C}_{i}$ and its depth information $\mathrm{D}_{i}$. 
The tracking process begins by distinguishing objects $\mathrm{O}_{i}$ from the background $\mathrm{B}_{i}$ through the instance masks $\mathrm{S}_{i}$ generated by YOLO segmentation \cite{7780460}.
The set $O_{i}$ may contain $N$ objects, each assigned an object ID $j$, with the $j^{th}$ object denoted as $\mathrm{O}_{i}(j)$ for $j \in [0, N]$.
Specifically, $O_{i}$ includes both static and dynamic objects, where the dynamic ones are easily misidentified and need to be filtered out. 
To minimize reliance on prior environmental knowledge, we initialize Gaussians $\mathrm{G}_i^{\mathrm{O}}$ and $\mathrm{G}_i^{\mathrm{B}}$ to represent both objects and background without requiring detailed semantics of their dynamic features. 
Subsequently, we render these Gaussians to compute the rendering loss flows using a photometric-geometric loss function. 
This process supports us in filtering out dynamic objects and optimizing the camera pose.  
The detailed procedure is outlined in the \enquote{Dynamic Object Prune} section, marked by a red dashed box. 
The implementation details are provided in the following section.

After filtering out dynamic objects using the loss flows, Gassidy computes an object-level joint loss for optimizing the camera pose. 
Subsequently, we determine the keyframe that exhibits significant changes compared to the previous one. 
Once a keyframe is selected, mapping proceeds by constructing the currently visible regions, while rendering loss in pruned areas is excluded from optimization. 
Pruned regions are reconstructed when subsequent keyframes provide sufficient data. 
Finally, the features of the Gaussians are updated based on the keyframe, and Gassidy iteratively repeats this process until all images are processed, resulting in a cleanly constructed scene without dynamic object disturbances.

\subsection{Scene representing using 3DGS}
\label{3b}
To handle unpredictable disturbances from dynamic objects, we utilize 3DGS to represent environmental changes as Gaussian variations across different regions, guided by instance segmentation for potential dynamic object \cite{kerbl20233d}.
To apply 3DGS, we first need to perform the Gaussian initialization. 
The first step of initialization involves converting $\mathrm{O}_{i}$ and $\mathrm{B}_{i}$ into object point clouds $\mathrm{P}_{i}^{\mathrm{O}(j)} \in \mathbb{R}^{a_j \times 7}$ and background point clouds $\mathrm{P}_i^{\mathrm{B}} \in \mathbb{R}^{b \times 7}$, where $\mathrm{a}_j$ is the number of points in the $j^{th}$ object, and $\mathrm{b}$ is the number of points in the background. 
The 7 channels of features in each point cloud consist of 3 RGB channels, 3 coordinate channels, and 1 object ID ($j$) channel. 
The coordinates and object ID in $\mathrm{P}_{i}^{\mathrm{O}(j)}$ and $\mathrm{P}_i^{\mathrm{B}}$ are then used to initialize the Gaussians $\mathrm{G}_i$. 
Following the method proposed in \cite{kerbl20233d}, our Gaussian function $\mathrm{G}_i(x)$ is: 
\begin{align}
    &\mathrm{G}_i(x) = \mathrm{e}^{-\frac{1}{2} (x - \mu_i)^T (\Sigma_i)^{-1} (x - \mu_i)}\text{,}
\end{align}
where $\mu_i$ is the mean vector representing the position of each Gaussian and $\Sigma_i$ is the full 3D covariance matrix that defines the spread and orientation of the Gaussian in world space.
At this stage, the initial size is set through the scale vector in $\Sigma_i$, which has identical values in all three directions, specifically the mean depth value (Z). 
Additionally, the orientation is initialized with an identity quaternion in $\Sigma_i$. 
Next, we synthesize color and opacity information from the point cloud by splatting and blending the Gaussians, as described in GS-SLAM \cite{Matsuki:Murai:etal:CVPR2024}. 
This process results in the initialization of Gaussians for objects $\mathrm{G}_{i}^{\mathrm{O}(j)}$ and the background $\mathrm{G}_i^{\mathrm{B}}$.

Subsequently, we perform rendering to compute the loss by projecting the 3D Gaussians ($\Sigma_{W}$, $\mu_{W}$) onto 2D space ($\Sigma_{I}$, $\mu_{I}$) for both objects and background, using the formula:
\begin{align}
   &\Sigma_{I,i}^{\mathrm{B}\cup \mathrm{O}} = \mathrm{J}_{i}^{\mathrm{B}\cup \mathrm{O}}\mathrm{R}_{i}\Sigma_{W,i}^{\mathrm{B}\cup \mathrm{O}}{\mathrm{R}_{i}}^T{\mathrm{J}_{i}^{\mathrm{B}\cup \mathrm{O}}}^T\text{,}
\end{align}
where $O$ and $B$ denote the sets of objects and background, $\Sigma_{I,i}^{O}$ denotes the 2D covariance matrix for the objects in the $i^{th}$ frame of the input RGB image, while $\Sigma_{I,i}^{B}$ corresponds to the 2D covariance matrix for the background in the same frame. 
Here, the Jacobian $\mathrm{J}_i^{\mathrm{B}\cup \mathrm{O}}$ approximates the projective transformation for each Gaussian in the background and object set, and $\mathrm{R}$ is the rotation matrix derived from the camera pose. 
The mean $\mu_I$ in the 2D space is computed as: 
\begin{align}
   &\mu_{I,i}^{\mathrm{B}\cup \mathrm{O}} = \pi(\mathrm{R}_{i}\mu_{W,i}^{\mathrm{B}\cup \mathrm{O}} + \mathrm{t}_i)\text{,}
\end{align}
where $\pi$ represents the projection operation, and $\mathrm{t}_i$ is the translation component of the camera pose. 
Next, the $\mathrm{C}_{i}$ and $\mathrm{D}_{i}$ (during the initialization, $i=1$) will be employed as ground truth and start the initial mapping process for enough iterations, updating the features ($\Sigma_i$ and color) of Gaussians. 
The loss employed in mapping will be detailed in the next section. 
After the initial mapping, we obtain a set of fully learned Guassians that can represent the scene through rendering. 
Finally, Gassidy proceeds with camera tracking and mapping for subsequent inputs by utilizing the method provided by GS-SLAM, which leverages the Jacobian to guide optimization and minimize the error between the estimated and observed data.

\begin{table*}[tb]
        \setlength{\tabcolsep}{2.5mm}
        \fontsize{7.5}{8}\selectfont
		\caption{Camera tracking results on dynamic scenes from the \textbf{TUM RGB-D} dataset. The best results within each domain are highlighted in \textbf{bold}, and the best results among all domains are marked with \underline{underline}. D/S indicates whether the method is dense or sparse reconstruction, \enquote{ATE} column shows the RMSE of the ATE, and the \enquote{Std.} column presents the standard deviation of ATE.
        X means tracking failure, and \textminus  \space indicates not mentioned in the original report. }
		\centering
		\label{table: TUM-rgbd}
		\sisetup{group-digits=false,table-number-alignment=right,round-mode=places,round-precision=1}%
	\begin{tabular}{S S | r r r r r r r r r r}
		\toprule
		{\textbf{Methods}}& {\textbf{Type}} & \multicolumn{2}{c}{{\textit{f3/wk\_xyz}}} & \multicolumn{2}{c}{{\textit{f3/wk\_hf}}} & \multicolumn{2}{c}{{\textit{f3/wk\_st}}} & \multicolumn{2}{c}{{\textit{f3/st\_hf}}} & \multicolumn{2}{c}{{\textbf{Avg.}}} \\
        \cmidrule(lr){1-1}
        \cmidrule(lr){2-2}
		\cmidrule(lr){3-4}
		\cmidrule(lr){5-6}
		\cmidrule(lr){7-8} \cmidrule(lr){9-10} \cmidrule(lr){11-12} 
  
		\textbf{Keypoint-based SLAM methods}  
            & \textit{D/S} & {ATE} & {Std.} & {ATE} & {Std.} & {ATE} & {Std.} & {ATE} & {Std.} & {ATE} & {Std.} \\
            
        {ORB-SLAM3} \cite{ORBSLAM3_2020} 
            & \textbf{S} & {28.1} & {12.2} & {30.5} & {9.0} & {2.0} & {1.1} & {\textbf{2.6}} & {1.6}  
            & 15.8 & 6.0 \\
            
        {DynaSLAM} \cite{8421015}  
            & \textbf{S} & \underline{\textbf{1.7}} & {\textminus} & {\underline{\textbf{2.6}}} & {\textminus} & {\textbf{0.7}} & {\textminus} & {2.8} & {\textminus} 
        & \underline{\textbf{2.0}} & {\textminus} \\
        
        \midrule
        
		\textbf{NeRF-based SLAM methods}  
            & \textit{D/S} & {ATE} & {Std.} & {ATE} & {Std.} & {ATE} & {Std.} & {ATE} & {Std.} & {ATE} & {Std.} \\
            
        {NICE-SLAM} \cite{NICE-SLAM} 
            & \textbf{D} & {113.8} & {42.9} & X & X & 137.3 & 21.7 & 93.3 & 35.3  
        & 114.8 & 33.3 \\
        
        
        {RoDyn-SLAM} \cite{jiang2024rodyn}  
            & \textbf{D}  & \textbf{8.3}     & \textbf{5.5}     & \textbf{5.6}    & \textbf{2.8}   & \textbf{1.7}           & \textbf{0.9}   & \textbf{4.4}  & \textbf{2.2} 
                & \textbf{4.1} & \textbf{2.3} \\
        
        \midrule
        \textbf{3DGS-based SLAM methods}  
            & \textit{D/S} & {ATE} & {Std.} & {ATE} & {Std.} & {ATE} & {Std.} & {ATE} & {Std.} & {ATE} & {Std.} \\
            
        {GS-SLAM} \cite{Matsuki:Murai:etal:CVPR2024} 
            & \textbf{D} & 37.2    & 9.9    & 60.0     & 20.7   & 8.4           & 4.1   & 7.4    & 5.4 
                & 28.5 & 10.0 \\
                
        {SplaTAM} \cite{keetha2024splatam}  
            & \textbf{D} & 149.2    & 37.4      & 157.8 & 54.4      & 85.3         & 16.1     & 14.0    & 6.8  
                & 125.6 & 109.6 \\
                
        {Gaussian-SLAM} \cite{yugay2023gaussianslam} 
            & \textbf{D} & 133.7    & 54.8   & 80.7   & 31.6   & 19.1         & 5.2   & 5.4    & 2.2 
                & 59.7 & 23.5 \\
                
        \textbf{GassiDy (Ours)} 
            & \textbf{D} & \textbf{3.5}     & \textbf{1.6}    & \textbf{3.7}    & \textbf{1.9}   & \underline{\textbf{0.6}} & \textbf{0.3}    & \underline{\textbf{2.4}}    & \textbf{1.4} 
                & \textbf{2.6} & \textbf{1.3} \\
                
		\bottomrule
	\end{tabular}
\end{table*}

\begin{table*}[tb]
        \setlength{\tabcolsep}{2mm}
        \fontsize{7.5}{8}\selectfont
		\caption{Camera tracking results on dynamic scenes from the \textbf{BONN Dynamic RGB-D} dataset. The notation is identical to Table \ref{table: TUM-rgbd}.}
		\centering
		\label{table: BONN Dynamic RGB-D}
		\sisetup{group-digits=false,table-number-alignment=right,round-mode=places,round-precision=1}%
	\begin{tabular}{S S | r r r r r r r r r r r r}
		\toprule
		{\textbf{Methods}}& {\textbf{Type}} & \multicolumn{2}{c}{{\textit{balloon}}} & \multicolumn{2}{c}{{\textit{balloon2}}} & \multicolumn{2}{c}{{\textit{ps\_track}}} & \multicolumn{2}{c}{{\textit{ps\_track2}}} & \multicolumn{2}{c}{{\textit{mv\_box2}}} & \multicolumn{2}{c}{{\textbf{Avg.}}} \\
        \cmidrule(lr){1-1}
        \cmidrule(lr){2-2}
		\cmidrule(lr){3-4}
		\cmidrule(lr){5-6}
		\cmidrule(lr){7-8} \cmidrule(lr){9-10} \cmidrule(lr){11-12} \cmidrule(lr){13-14} 
		\textbf{Keypoint-based SLAM methods}  & \textit{D/S} & {ATE} & {Std.} & {ATE} & {Std.} & {ATE} & {Std.} & {ATE} & {Std.} & {ATE} & {Std.} & {ATE} & {Std.} \\
        {ORB-SLAM3} \cite{ORBSLAM3_2020}  & \textbf{S} & {5.8} & {2.8} & {17.7} & {8.6} & {70.7} & {32.6} & {77.9} & {43.8} 
        & \underline{\textbf{3.1}} & {1.6} & {35.0} & {17.9} \\
        {DynaSLAM} \cite{8421015}  & \textbf{S} & \textbf{3.0} & \textminus & \underline{\textbf{2.9}} & \textminus & \underline{\textbf{6.1}} & \textminus & \underline{\textbf{7.8}} & \textminus 
        & {3.9} & \textminus & \underline{\textbf{4.74}} & \textminus \\
        \midrule
		\textbf{NeRF-based SLAM methods}  & \textit{D/S} & {ATE} & {Std.} & {ATE} & {Std.} & {ATE} & {Std.} & {ATE} & {Std.} & {ATE} & {Std.} & {ATE} & {Std.}\\
        {NICE-SLAM} \cite{NICE-SLAM}  & \textbf{D} & X & X & {66.8} & {20.0} & {54.9} & {27.5} & {45.3} & {17.5} 
        & {31.9} & {13.6} & {49.7} & {19.7}  \\
        {RoDyn-SLAM}\cite{jiang2024rodyn}  & \textbf{D} & \textbf{7.9} & \textbf{2.7} & \textbf{11.5} & \textbf{6.1} & \textbf{14.5} & \textbf{4.6} & \textbf{13.8} & \textbf{3.5} 
        & \textbf{12.6} & \textbf{4.7} & \textbf{12.1} & \textbf{4.32} \\
        \midrule
        \textbf{3DGS-based SLAM methods}  & \textit{D/S} & {ATE} & {Std.} & {ATE} & {Std.} & {ATE} & {Std.} & {ATE} & {Std.} & {ATE} & {Std.} & {ATE} & {Std.} \\
        {GS-SLAM} \cite{Matsuki:Murai:etal:CVPR2024}  & \textbf{D} & {39.5} & {19.3} & {35.6} & {19.5} & {93.3} & {36.3} & {51.2} & {19.1} 
        & {6.1} & {4.5} & {45.1} & {19.7} \\
        {SplaTAM }\cite{keetha2024splatam}  & \textbf{D} & {35.8} & {14.3} & {38.7} & {15.0} & {138.4} & {48.1} & {126.3} & {36.7} 
        &{22.0}& {12.3} & {54.6} & 33.7 \\
        {Gaussian-SLAM} \cite{yugay2023gaussianslam}  & \textbf{D} & {65.2} & {25.5} & {34.8} & {22.1} & {109.2} & {58.9} & {118.7} & {57.2} 
        & {31.7} & {20.9} & {71.9} & {36.9} \\
        \textbf{GassiDy (Ours)}  & \textbf{D} & \underline{\textbf{2.6}} & \textbf{0.8} & \textbf{7.6} & \textbf{3.4} & \textbf{10.3} & \textbf{4.4} & \textbf{13.0} & \textbf{4.8} 
        & \textbf{5.4} & \textbf{1.9} & \textbf{7.8} & \textbf{3.1} \\
		\bottomrule
	\end{tabular}
	\vspace{-1.5em}
\end{table*}

\subsection{Loss define and dynamic objects identification}
\label{3c}
In this section, we define the loss that will be utilized in tracking and mapping optimization under dynamic environments as well as introduce the logic of dynamic object filtering. 
Unlike the other 3DGS SLAM, our approach segments the scene into objects of interest and background, utilizing errors from both to optimize the camera pose and mapping accuracy. 
The loss utilized for optimization in $i^{th}$ frame is:
\begin{subequations}
\begin{align}
   &\mathrm{L}_\text{pho}^{\mathrm{O}(j)} = \frac{1}{\mathrm{a}_j}\sum_{p \in \mathrm{O}(j)}{(|\hat{\mathrm{I}}_p - \mathrm{I}_p| \circ \mathrm{S}_{O(j)})}\text{,}\\
   &\mathrm{L}_\text{pho}^{\mathrm{B}} = \frac{1}{\mathrm{b}}\sum_{p \in \mathrm{B}}{(|\hat{\mathrm{I}}_p - \mathrm{I}_p| \circ \neg \bigcup_{\mathrm{O}(j)\in O} \mathrm{S}_{O(j)})}\text{,}
\end{align}
\end{subequations}
where $\mathrm{L}_\text{pho}^{\mathrm{O}(j)}$ and $\mathrm{L}_\text{pho}^{\mathrm{B}}$ represent the mean photometric rendering loss for the $j^{th}$ object and the background in this frame, respectively, $p$ denote the pixles in the objects or background, $\hat{\mathrm{I}}$ is the predicted image through rendering among Gaussians, $\mathrm{I}$ is the ground truth image, $\mathrm{S}_{\mathrm{O}(j)}$ is the mask for the $j^{th}$ object in the frame, and $\circ$ indicates that the mask $\mathrm{S}_{\mathrm{O}(j)}$ is applied to the loss map. 
The geometric rendering loss $\mathrm{L}_\text{geo}^{\mathrm{O}(j)}$ and $\mathrm{L}_\text{geo}^{\mathrm{B}}$ are calculated via the identical approach. 
Additionally, we utilize datasets containing real-world data, where depth values can be unreliable due to inconsistent lighting and sudden changes in highly dynamic environments. 
To address this challenge, we developed an adaptive loss function that dynamically adjusts the weighting between photometric and geometric rendering losses during tracking and mapping:
\begin{align}
    &\mathrm{L} = \lambda_{a} \mathrm{L}_{pho} + (1-\lambda_{a})\mathrm{L}_{geo}
    \label{formulaloss}\text{,}
\end{align}
where $\mathrm{L}$ represents the combined rendering loss used for optimization, $\mathrm{L}_{pho}$ and $\mathrm{L}_{geo}$ denote the photometric and geometric rendering losses, respectively, and $\lambda{a}$ is an adaptive weight that varies between $[\lambda_\text{lo},\lambda_\text{up}]$ depending on the quality of the depth map. 
The quality is assessed by the proportion of zero or NaN values within the depth map. As the quality decreases, $\lambda_{a}$ increases; we currently model this relationship using a linear function.

We leverage the rendering loss defined in formula~\ref{formulaloss} for both optimization and dynamic object filtering. 
To filter out dynamic objects while avoiding the misidentification of subtle object movements, we apply a coarse-to-fine strategy. 
Initially, we perform 40 iterations of loss computation for both objects and the background, coarsely optimizing the camera pose using only $\mathrm{L}_\text{pho}^{\mathrm{B}}$ and $\mathrm{L}_\text{geo}^{\mathrm{B}}$, while updating the object loss at each iteration. 
To track how the loss evolves for static and dynamic objects, we calculate the loss difference between the $k^{th}$ and $(k+1)^{th}$ iterations for $k \in [1,40]$. 
After gathering the rendering loss flows over the 40 iterations, we apply a Gaussian Mixture Model (GMM) to classify the background and objects. 
The key insight is that the loss for the background and static objects decreases consistently over iterations as they become well-aligned with the scene geometry. 
In contrast, dynamic objects exhibit higher and more fluctuating loss values across iterations due to their motion as shown in Fig.~\ref{fig:workflow}. 
Based on this amplified distinction, we apply the following rule to prune dynamic objects:
\begin{align}
   &\mathrm{P}(\mathrm{O}(j) \in Dynamic | \Delta \mathrm{L}^{\mathrm{O}(j)}) > \theta \implies \mathrm{O}_d \cup \mathrm{O}(j)\text{,} 
\end{align}
where $\mathrm{P}$ is the possibility generated by GMM that $\mathrm{O}(j)$ is dynamic, $\Delta \mathrm{L}^{\mathrm{O}(j)}$ represent the rendering loss flow calculated during iteration for the $j^{th}$ object, the threshold $\theta$ determines whether an object should be treated as dynamic, with $\mathrm{O}_d$ serving as a set to store dynamic objects. 
Moreover, we maintain a hash table to manage the life cycle of dynamic objects. 
For example, if an object is initially classified as static in a single frame, it will not be reintroduced for optimization. 
Only after being consistently classified as static across three consecutive frames will its information be added back for optimization. 
Subsequently, all Gaussians with IDs matching those of the objects in $\mathrm{O}_d$ are pruned if the last frame is selected as a keyframe. 
The final object level joint loss $\mathrm{L}_t$ used for fine camera pose optimization until it converges is computed as:
\begin{align}
    \mathrm{L}_t = \mathrm{L}^\mathrm{B} + \sum_{\mathrm{O}(j) \in \mathrm{O}_e}\mathrm{L}^{\mathrm{O}(j)}\text{,}
\end{align}
where $O_e = O_d \setminus O$. 

\begin{figure}[tb]
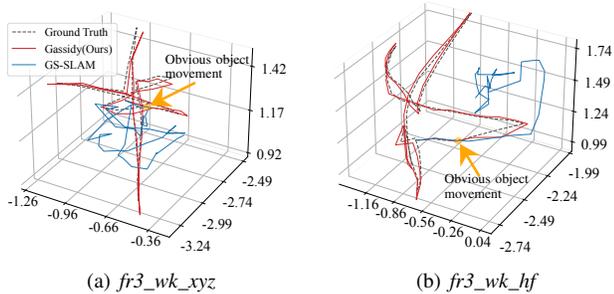

    \centering
    \subfigure[\textit{fr3\_wk\_xyz}]{
            \input{tikz_files/traj_1}
		\label{fig:tra_1}
	}
 \hfill
    \subfigure[\textit{fr3\_wk\_hf}]{
            \input{tikz_files/traj_2}
		\label{fig:tra_2}
	}
    \vspace{-0.5em}
    \caption{Camera tracking trajectories of Gassidy and GS-SLAM in dynamic scenes from TUM dataset. Each method's trajectory, along with the ground truth, is highlighted using different colors, with the moment of significant object movement indicated by an arrow.}
    \label{Camera tracking}
    \vspace{-0.8em}
\end{figure}

\begin{figure}[tb]
    \centering
    \input{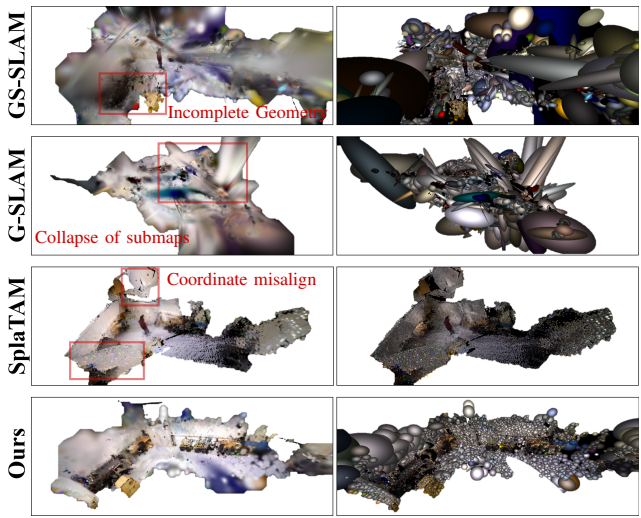}
    \vspace{-1.7em}
    \caption{Large scene reconstruction quality comparison between our method and other 3DGS-based methods in \textit{person\_track} scene from BONN dataset. The red box highlights the flaws in these methods.} 
    \label{fig:large_comparison2}
    \vspace{-1.5em}
\end{figure}

After filtering out dynamic objects and accurately estimating the camera pose for the input image, we need to identify the keyframe that provides sufficient new information to update the Gaussians during mapping. 
Keyframe selection is based on co-visibility checks using Intersection over Union (IoU) and Overlap Coefficient (OC). 
They are chosen if their IoU is below 80\% and their OC exceeds 20\%, ensuring comprehensive coverage while filtering out frames with excessive changes. 
Once the keyframe is determined, we compute the photometric-geometric $\mathrm{L}_m$ loss for mapping optimization:
\begin{align}
  \mathrm{L}_m = \text{Mean}((\hat{\mathrm{I}} - \mathrm{I}) \circ \neg \mathrm{S}_{\mathrm{O}_d})  \text{,}
\end{align}
where $\mathrm{S}_{\mathrm{O}_d}$ contains the masks for dynamic objects, managed utilizing the lifecycle method described above. 
It is updated by referencing the maintained hash table and including dynamic objects still present in the scene. 
This approach ensures that the loss from dynamic objects does not affect the mapping quality of static components. 
As illustrated in the section highlighted with a green dashed box in Fig.~\ref{fig:workflow}, the Gaussians used for mapping may not cover areas with dynamic objects due to our filtering process during camera tracking. 
The mapping process will focuses on constructing the visible static scene, with Gaussians being added or adjusted when subsequent keyframes provide information about previously empty areas.


\section{EXPERIMENT}

\begin{figure}[tb]
    \centering
    \includegraphics[width=0.48\textwidth]{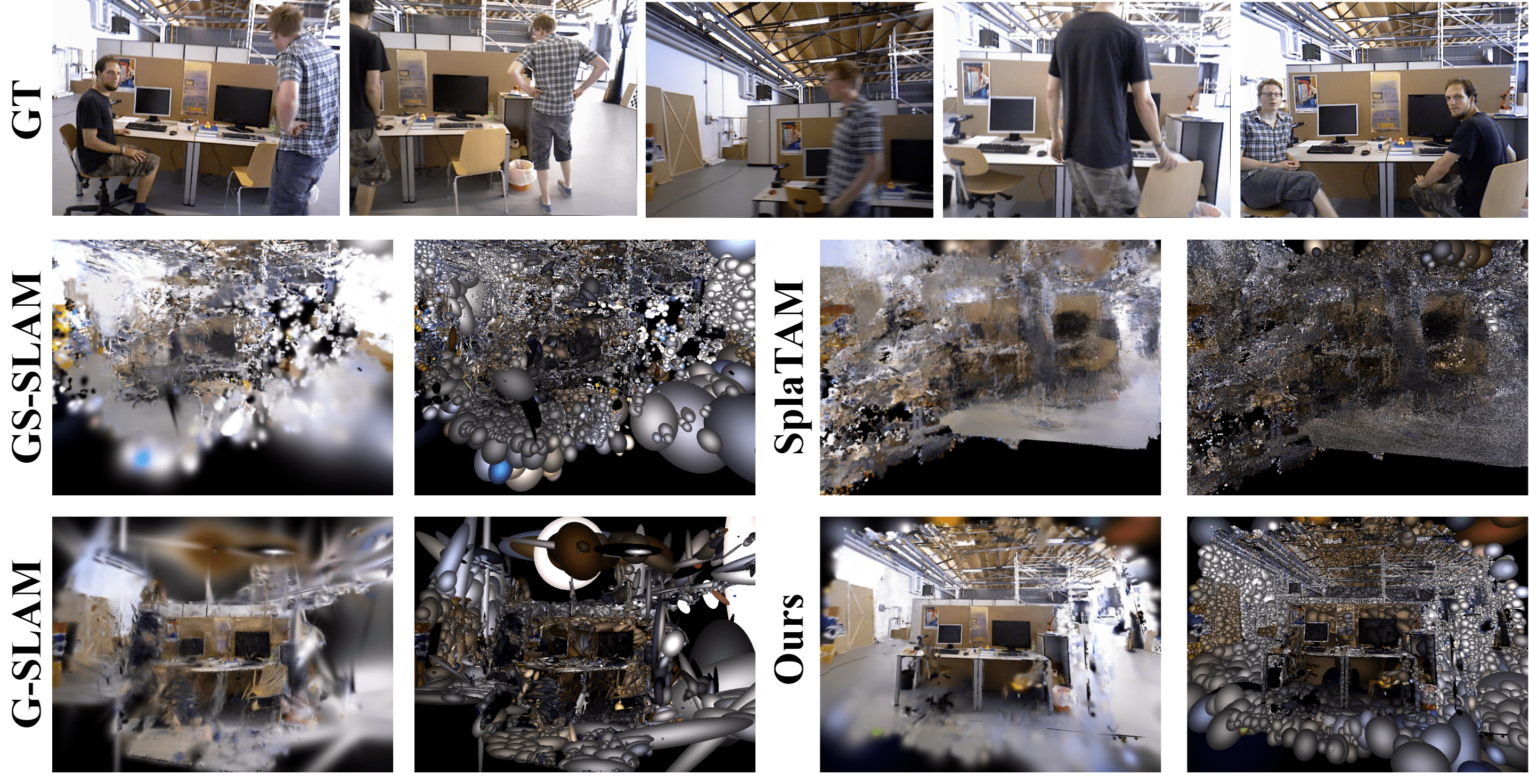}
    \caption{Map reconstruction quality comparison between our method and other 3DGS-based methods on the \textit{fr3/walk\_xyz} scene from TUM dataset. The image on the left shows the rendered RGB, while the image on the right is the generated Gaussian map.}
    \label{fig:map_comparison1}
    \vspace{-1.5em}
\end{figure}

To evaluate the proposed Gassidy, we conduct extensive comparison experiments against state-of-the-art SLAM methods. We leverage a variety of metrics to assess camera tracking accuracy and map quality.
\subsection{Experiment Setup}
We selected a diverse range of systems, including well-known SLAM methods for static environments \cite{ORBSLAM3_2020, NICE-SLAM,Matsuki:Murai:etal:CVPR2024,keetha2024splatam,yugay2023gaussianslam} and methods optimized for dynamic environments \cite{8421015,jiang2024rodyn}. 
As no 3DGS-based SLAM methods have been developed for dynamic environments, only those designed for static environments are included.
Our experiments were conducted on two real-world public datasets: TUM RGB-D \cite{sturm12iros} and BONN RGB-D Dynamic \cite{palazzolo2019arxiv}. 
We selected a variety of dynamic scenes from both datasets, including \textit{fr3/wk\_xyz}, \textit{fr3/wk\_hf}, \textit{fr3/wk\_st}, and \textit{fr3/st\_hf} from TUM, as well as \textit{balloon}, \textit{balloon2}, \textit{ps\_track}, \textit{ps\_track2}, and \textit{mv\_box2} from the BONN Dynamic dataset. 
It is important to note that our benchmarks focus exclusively on 3DGS-based SLAM methods, while results for other domains are sourced from their respective publications. 
All experiments were conducted on a computer with an Intel i9-12900K CPU and an NVIDIA RTX3080 GPU. 
The GMM configuration employs an adaptive approach to determine the appropriate number of components (\textit{n\_components}). 
Specifically, if the \textit{AIC} value exceeds 0, the number of components is reduced to 1; otherwise, \textit{n\_components} is set to 2. 
This method effectively mitigates overfitting of GMM, which in this context refers to the misclassification of static objects as dynamic.
The adaptive ratio range $[\lambda_\text{lo},\lambda_\text{up}]$ of photometric-geometric loss is in $[0.88,0.95]$. 
The threshold $\theta$ is set to 99.9\%. 
For point cloud down-sampling, we use a factor of 128/32 for the TUM RGB-D dataset and 256/64 for the BONN dynamic dataset.

For evaluating camera tracking precision, we use the root mean square error (RMSE) and standard deviation (Std.) of the absolute trajectory error (ATE). 
For map quality evaluation, we employ standard photometric rendering metrics: peak signal-to-noise ratio (PSNR), which indicates image clarity, and structural similarity index measure (SSIM) along with learned perceptual image patch similarity (LPIPS), both of which evaluate the similarity between the input and rendered images. 
The Map quality metrics are recorded every five frames for detailed analysis.

\begin{table}[tb]
    \setlength{\tabcolsep}{1.8mm}
    \fontsize{7.5}{8}\selectfont
	\caption{Map quality results on dynamic scenes in \textbf{BONN-RGBD} dataset. The units are [dB,\%,\%] for PSNR, SSIM, and LPIPS, respectively. The \colorbox{ForestGreen}{\textbf{best}}, \colorbox{MustardYellow}{second-best}, and \colorbox{BurntOrange}{third-best} value are highlighted in different colors.}
	\centering
	\label{table:bonn-results-vertical}
	\sisetup{table-number-alignment=right,table-parse-only=true}%
	\begin{tabular}{cccccc}
	   \toprule
		{\textbf{Metrics}} & {\textbf{Scene}} & {\makecell{GS-SLAM \\ \cite{Matsuki:Murai:etal:CVPR2024}}} & {\makecell{SplaTam \\ \cite{keetha2024splatam}}} & {\makecell{G-SLAM \\ \cite{palazzolo2019arxiv}}} & {\makecell[c]{Gassidy\\(Ours)}}  \\
	   \midrule
		
        \multirow{5}{*}{$\textbf{PSNR}_\uparrow$} 
        &   \textit{balloon}   &  \cellcolor{BurntOrange}17.7 & 17.6 & \cellcolor{MustardYellow}21.6 & \cellcolor{ForestGreen}\textbf{24.0}    \\
        
        & \textit{balloon2}   & \cellcolor{BurntOrange}19.4 & 16.8 &\cellcolor{MustardYellow}19.8 & \cellcolor{ForestGreen}\textbf{22.9}    \\
        
        & \textit{ps\_track}   & \cellcolor{BurntOrange}18.9 & 18.8 & \cellcolor{MustardYellow}24.0 & \cellcolor{ForestGreen}\textbf{24.6}    \\

        & \textit{ps\_track2}   & \cellcolor{BurntOrange}20.0 & 17.5 & \cellcolor{MustardYellow}23.6 & \cellcolor{ForestGreen}\textbf{24.2}    \\

        & \textit{mv\_box2}    & \cellcolor{BurntOrange}23.5 & 20.2 & \cellcolor{MustardYellow}25.1 & \cellcolor{ForestGreen}\textbf{25.5}    \\
     \midrule
        \multirow{5}{*}{$\textbf{SSIM}_\uparrow$} 
        &  \textit{balloon}   & 71.4 & \cellcolor{BurntOrange}76.9 & \cellcolor{ForestGreen}\textbf{84.7} & \cellcolor{MustardYellow}77.5    \\
        
        & \textit{balloon2}   & \cellcolor{MustardYellow}74.8 & 64.4 & \cellcolor{ForestGreen}\textbf{77.4} & \cellcolor{BurntOrange}71.5    \\

        &   \textit{ps\_track}   & \cellcolor{BurntOrange}73.1 & 68.8 & \cellcolor{ForestGreen}\textbf{90.6} & \cellcolor{MustardYellow}78.7    \\

         &  \textit{ps\_track2}    & \cellcolor{BurntOrange}75.6 & 71.8 & \cellcolor{ForestGreen}\textbf{89.8} & 
         \cellcolor{MustardYellow}77.3    \\
        
         & \textit{mv\_box2}    & 83.6 & \cellcolor{BurntOrange}82.7 & \cellcolor{ForestGreen}\textbf{89.4} & \cellcolor{MustardYellow}85.0    \\
        
	   \midrule   
     \midrule
        \multirow{5}{*}{$\textbf{LPIPS}_\downarrow$}   
	    &  \textit{balloon} & 48.0 & \cellcolor{ForestGreen}\textbf{24.3} & \cellcolor{MustardYellow}27.5 & \cellcolor{BurntOrange}32.5    \\

        &  \textit{balloon2} & \cellcolor{BurntOrange}36.6 & \cellcolor{ForestGreen}\textbf{32.6} & \cellcolor{MustardYellow}35.6 & 39.4    \\
        
	    &  \textit{ps\_track} & 39.6 & \cellcolor{MustardYellow}27.5 & \cellcolor{ForestGreen}\textbf{20.5} & \cellcolor{BurntOrange}32.8    \\ 

        & \textit{ps\_track2}  & 37.8 & \cellcolor{MustardYellow}26.3 & \cellcolor{ForestGreen}\textbf{20.2} & \cellcolor{BurntOrange}32.0    \\

        & \textit{mv\_box2}  & 26.1 & \cellcolor{ForestGreen}\textbf{18.7} & \cellcolor{MustardYellow}22.3 & \cellcolor{BurntOrange}25.1    \\    
		\bottomrule
	\end{tabular}
	\vspace{-1.5em}
\end{table}
\subsection{Camera Tracking Precision Analysis}
The results for the TUM dataset are presented in Table~\ref{table: TUM-rgbd}. 
Compared to GS-SLAM, SplaTAM, and Gaussian-SLAM, Gassidy improves the RMSE ATE by an average of 90.9\%, 97.9\%, and 95.6\%, respectively, and enhances the standard deviation by 87.0\%, 98.8\%, and 94.5\%. 
As a result, we consistently achieve the best performance among 3DGS-based methods. 
That is because, in those approaches, dynamic objects may be regarded as static objects influenced by slight camera movement, thereby reducing tracking accuracy. 
In contrast, Gassidy can amplify the difference between static and dynamic objects by iteratively analyzing the features of static and dynamic objects, thereby accurately filtering them. 
In terms of the NeRF-based NICE-SLAM and RoDyn-SLAM, Gassidy shows an average improvement of 97.7\% and 36.6\% in RMSE ATE, respectively, while also reducing the standard deviation by 96.1\% and 43.5\%.  
This is because these approaches depend on detailed semantic segmentation based on prior knowledge. 
Consequently, they cannot accurately filter out unpredictable objects that are not included in the prior semantics.  
In contrast, 3DGS can leverage a larger number of input samples for rendering loss computation and optimization, resulting in more detailed information and enhanced performance. 
Against sparse SLAM methods ORB-SLAM3, Gassidy delivers an average improvement of 83.5\% in RMSE ATE and 78.3\% in standard deviation. 
Compared to DynaSLAM, Gassidy improves RMSE ATE by 14.3\% in the \textit{f3/wk\_st} and \textit{f3/st\_hf} scenes.
And DynaSLAM shows 25.9\% better performance in \textit{f3/wk\_xyz} and \textit{f3/wk\_hf}.
It is noteworthy that although DynaSLAM may attain the highest tracking performance, it consistently neglects environmental details during the mapping process. 

The results for the BONN dataset are presented in Table~\ref{table: BONN Dynamic RGB-D}. 
The conclusion drawn in the BONN dataset is similar to TUM.  
Our method substantially outperforms other 3DGS-based methods, achieving improvements of 82.7\% in RMSE and 84.3\% in standard deviation. 
When compared to the NeRF-based domain, our method delivers much better performance, with an average improvement of 35.5\%.  
Against DynaSLAM, our method still delivers comparable performance, showing better performance (13.3\%) in the \textit{balloon} scene, demonstrating its advanced tracking capabilities.  
To understand the reasons behind our performance, we visualize some experimental results. 
As indicated by the red box in Fig.~\ref{fig:large_comparison2}, the constructed scenes of compared methods are fragmented. 
This fragmentation results from overfitting during mapping, causing misalignment with the target coordinate system. 
In terms of Gassidy, we effectively capture and filter dynamic objects, enabling consistent alignment with the target coordinate system and enhancing camera tracking.  
Moreover, as shown in Tables~\ref{table: TUM-rgbd} and~\ref{table: BONN Dynamic RGB-D}, Gassidy consistently achieves the lowest standard deviation, demonstrating its stability. 
For a detailed analysis, we present the trajectories for \textit{fr3/wk\_xyz} and \textit{fr3/wk\_hf} as shown in Fig.~\ref{Camera tracking}.
Comparison between Gassidy and GS-SLAM (the second-best method) shows that both methods experience reduced tracking accuracy after dynamic object movements. 
In this case, GS-SLAM struggles to recover due to disturbances from the dynamic objects. 
In contrast, Gassidy is able to quickly realigns with the ground truth by filtering out dynamic objects, thereby improving the camera tracking precision.


\begin{figure}[tb]
    \centering
    \includegraphics[width=0.48\textwidth]{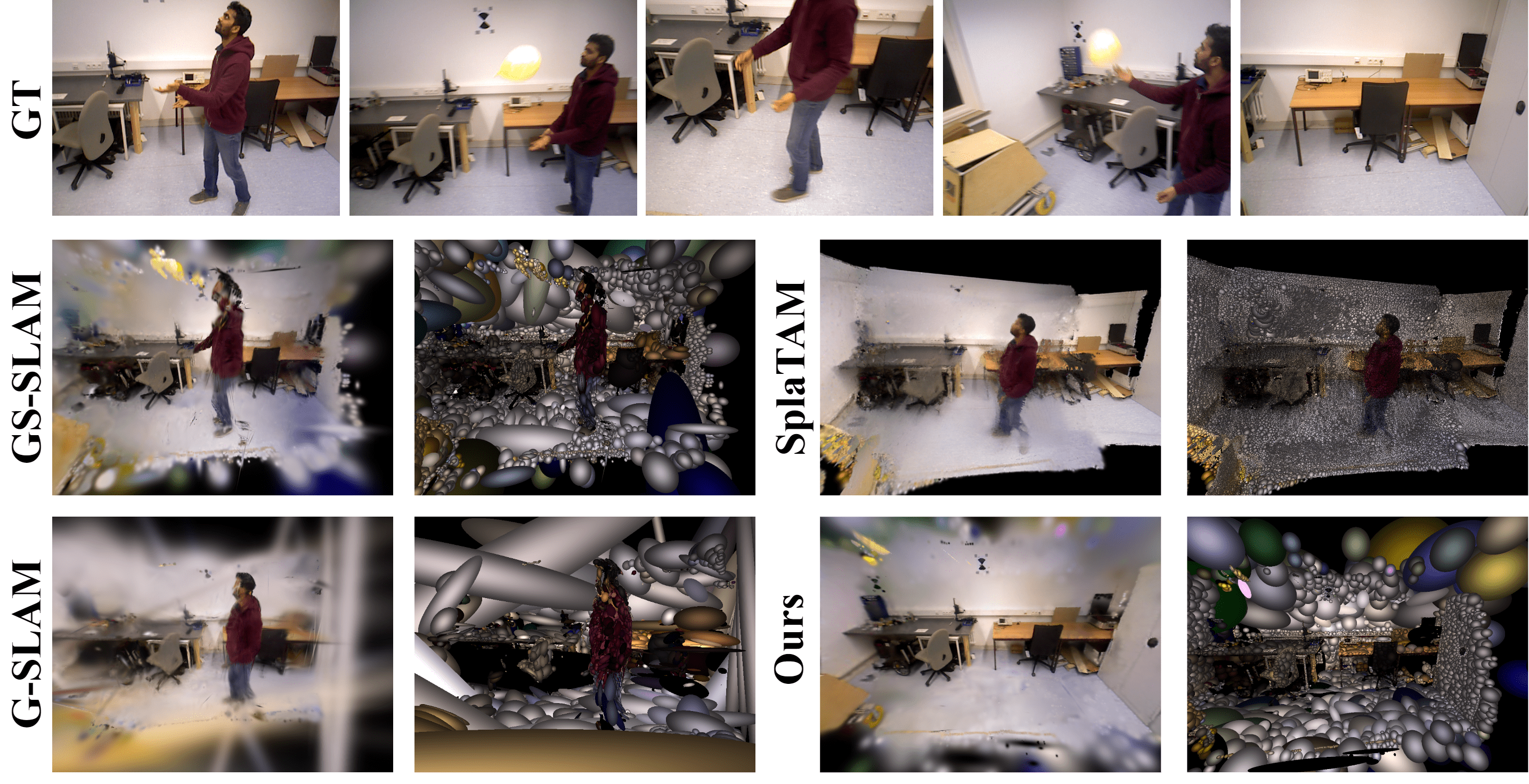}
    \caption{Map reconstruction quality comparison between our method and other 3DGS-based methods on the \textit{balloon} scene from BONN dataset. The image on the left shows the rendered RGB, while the image on the right is the Gaussian map generated by each method.}
    \label{fig:map_comparison2}
    \vspace{-1.5em}
\end{figure}

\subsection{Map Quality Analysis}
In Table~\ref{table:bonn-results-vertical}, the best, second-best, and third-best values are highlighted in distinct colors. 
It is important to note that the datasets employed lack clear ground truth baselines, so the comparison is based on input images and the results produced by different methods. 
In this context, as the SSIM and LPIPS reflect the similarity between the input and rendered images, approaches that can filter dynamic objects will result in lower values of these metrics. 
As illustrated in Figs.~\ref{fig:large_comparison2},~\ref{fig:map_comparison1}, and~\ref{fig:map_comparison2}, Gassidy, which filters dynamic objects, demonstrates better performance despite lower SSIM and LPIPS values. 
Moreover, for datasets containing both static and dynamic environments, our approach still achieves acceptable SSIM and LPIPS scores under the influence of the filtered dynamic objects. 
Furthermore, our method consistently delivers the best PSNR performance, with a 6.0\% average improvement over the second-best method. 
This improvement stems from the fact that disturbances caused by dynamic objects can lead to overfitting during scene reconstruction, resulting in a noisy map and inaccurate camera tracking.

\section{Conclusions}
We develop a dense RGB-D SLAM method called Gassidy which leverages a 3D Gaussian representation to handle dynamic environments effectively. 
To handle the disturbance from irregularly moving objects, we calculate the rendering loss flows for each environment component. 
By analyzing the loss change features in rendering loss flows, Gassidy distinguishes and filters out the dynamic objects, constructing a high-quality scene with accurate camera tracking. 
Moreover, our method reduces reliance on semantic priors by requiring only instance segmentation of potential dynamic objects, without needing prior knowledge of their dynamic features.
Our future work will focus on enhancing object-level reconstruction and the efficiency of the method for real-time robotics applications.

\bibliographystyle{IEEEtran}
\bibliography{icra2025}
\balance

\end{document}